%% file: ROSE_2024_File.tex
\documentclass[conference]{IEEEtran}

\usepackage[backend=biber,style=ieee]{biblatex} 
\ifCLASSINFOpdf
  \usepackage[pdftex]{graphicx}
  \graphicspath{{./figures/}}
  \DeclareGraphicsExtensions{.pdf,.jpeg,.png}
\else
\fi

\usepackage{tikz}
\usetikzlibrary{calc,positioning,shapes.geometric}
\usepackage{hhline}
\usepackage{colortbl}
\usepackage{csquotes}
\usepackage[inline]{enumitem}
\usepackage{url}
\usepackage{hyperref}
\usepackage{cleveref}
\usepackage[colorinlistoftodos]{todonotes}
\usepackage{amssymb}
\usepackage{color}
\usepackage{siunitx}
\usepackage{changes}
\usepackage{subfigure}
\usepackage{orcidlink}

\usepackage[nodayofweek,level]{datetime}

\definecolor{ColorSZ}{RGB}{190,174,255}
\definecolor{ColorGJ}{RGB}{204, 153, 0}
\definecolor{ColorNMS}{RGB}{170,255,255}

\newcommand{\martin}[1]{\todo[inline, color=ColorSZ!20, bordercolor=ColorSZ!0]{\scriptsize\textbf{Martin:} #1}}

\newcommand{\Cmpr}[1]{cf.~\Cref{#1}}

\begin{document}
\title{Using Unsupervised Learning to Explore Robot-Pedestrian Interactions in Urban Environments}

%\author{\IEEEauthorblockN{Sebastian Zug, Georg Jäger, Norman Seyffer, \\ Martin Plank, and Gero Licht}
%\IEEEauthorblockA{}
%\and
%\IEEEauthorblockN{Felix Wilhelm Siebert}
%\IEEEauthorblockA{Technical University of Denmark\\Dept. of Techn. Management and Economics\\ Lyngby, Denmark\\
%Email: \textit{felix@dtu.dk}}
%}

\author{\IEEEauthorblockN{Sebastian Zug, Georg Jäger\orcidlink{0000-0002-1356-2745}, Norman Seyffer \orcidlink{0009-0009-8039-6765}, \\ Martin Plank \orcidlink{0009-0004-6830-0293}, and Gero Licht}
\IEEEauthorblockA{\textit{TU Bergakademie Freiberg}\\
Freiberg, Germany \\
email: \{sebastian.zug, georg.jaeger, norman.seyffer, \\martin.plank, gero.licht\}@informatik.tu-freiberg.de}
\and
\IEEEauthorblockN{Felix Wilhelm Siebert \\ \textcolor{white}{irgendwas}}
\IEEEauthorblockA{\textit{Technical University of Denmark}\\
Lyngby, Denmark \\
email: felix@dtu.dk}
}


%Telephone: (800) 555--1212, Fax: (888) 555--1212}


\maketitle

\begin{abstract}

This study identifies a gap in data-driven approaches to robot-centric pedestrian interactions and proposes a corresponding pipeline.
The pipeline utilizes unsupervised learning techniques to identify patterns in interaction data of urban environments, specifically focusing on conflict scenarios. 
Analyzed features include the robot's and pedestrian's speed and contextual parameters such as proximity to intersections. 
They are extracted and reduced in dimensionality using Principal Component Analysis (PCA).
Finally, K-means clustering is employed to uncover underlying patterns in the interaction data.
A use case application of the pipeline is presented, utilizing real-world robot mission data from a mid-sized German city. 
The results indicate the need for enriching interaction representations with contextual information to enable fine-grained analysis and reasoning.
Nevertheless, they also highlight the need for expanding the data set and incorporating additional contextual factors to enhance the robots situational awareness and interaction quality.
\end{abstract}


% \fi
\IEEEpeerreviewmaketitle

%\todo[inline]{Add acknowledgments for R4R} Erledigt, ist als separater Abschnitt eingefügt.

\input{sections/00_introduction.tex}

\input{sections/01_soa.tex}

\input{sections/02_concept.tex}

\input{sections/03_implementation.tex}
\input{sections/04_future_work.tex}
\section*{Acknowledgment}

This work was partially founded by the German Federal Ministry for Digital and Transport within the mFUND program (grant no. 19FS2025A - Project \textit{Ready for Smart City Robots} and 19F1117 - Project \textit{RoboTraces}).

Johannes Kohl, Georg Muck, and Nico Zumpe supervised robot missions in 2023, which forms the basis for this work.

%\bibliographystyle{IEEEtran}
%\bibliography{ROSE_2024_File}
\printbibliography

\end{document}

%% file: sections/00_introduction.tex
\section{Motivation}

% The idea of small-sized robots operating on sidewalks and bicycle lane looks promising in many ways. In the currently most discussed area of application, last-mile delivery, robots provide very customer-oriented services, are more cost-effective and more environmentally friendly than a delivery van.~\cite{Figliozzi.2020}.
% In parallel studies and proof of concept implementations predict an additional increased market for mobile platforms collecting environmental data, accompanying  elderly people~\cite{Li.2023} or supporting workers in large outdoor areas~\cite{Marsden.2018}. With such applications, mobile robots become a new competitor for space resources in city centers. 
% % mit dem "concentrated" möchte ich ausdrücken, dass es ein "Hauptoperationsgebiet" gibt, in dem die Roboter sich bewegen. Darüber hinaus müssen dann aber auch andere Wegtypen genutzt werden, um zum Beispiel von der Straße zur Haustür zu kommen. 
% Should they be concentrated on pedestrian paths, bicycle tracks, on streets, or even on a separate \textquote{robot lane}?

%Small-sized robots offer promising applications in future smart cities. 
%They present a customer-centric, cost-effective, and environmentally friendly alternative for last-mile delivery tasks (e.g. ~\cite{Figliozzi.2020, Alverhed.2024, jennings2019study, jennings2020study, Figliozzi.2020}). 
%Pilot programs also explore their potential for environmental data collection~\cite{BAYAT201776}, assisting the elderly~\cite{Li.2023}, and supporting outdoor workers~\cite{Marsden.2018}.
Small-sized robots offer promising applications in future smart cities, providing a customer-centric, cost-effective, and environmentally friendly alternative for last-mile delivery tasks (e.g.~\cite{Figliozzi.2020, Alverhed.2024, jennings2019study, jennings2020study, Figliozzi.2020}). Pilot programs are also exploring their potential for environmental data collection~\cite{BAYAT201776}, assisting the elderly~\cite{Li.2023}, and supporting outdoor workers~\cite{Marsden.2018}. As these robots frequently interact with pedestrians and cyclists, ensuring safe and intuitive interactions is critical for social acceptance and technology adoption~\cite{abrams.2021, Arntz.2023}. 

% The deployment of small-sized robots holds promise for various applications in future smart cities. 
% In the popular domain of last-mile delivery, robots offer a customer-centric, cost-effective, and environmentally friendly alternative to delivery vans~\cite{Figliozzi.2020, Alverhed.2024, jennings2019study, jennings2020study, figliozzi2020carbon}.  
% Emerging pilot programs also show interests in platforms that collect environmental data~\cite{BAYAT201776}, assist elderly individuals~\cite{Li.2023}, or support workers in outdoor environments~\cite{Marsden.2018}. 
% These applications position mobile robots as new contenders for space in urban infrastructures like sidewalks, bicycle lanes and streets.
% As a result, they will frequently interact with pedestrians and cyclists. 
% These interactions need to be safe -- collision-free -- and intuitive to ensure social acceptance, which is crucial to the adoption of this technology \cite{abrams.2021, Arntz.2023}. 
%This requires the robots to interpret scenarios accurately.

% \Cref{fig:roboter_haustuer} illustrates the challenge.
% The robot, \textquote{Claudi}, is navigating on a sidewalk, only $\qty{1.2}{\meter}$ wide, further restricted by a truck on the right, when a pedestrian exits a building on the left.
% While the sidewalk's width is sufficient for the robot to operate, it is not sufficient for conflict-free passage.
% Such conflicts need to be identified and resolved to ensure safe and socially acceptable interactions between robots and passersby.

\begin{figure}
    \centering
    \includegraphics[width=0.8\linewidth]{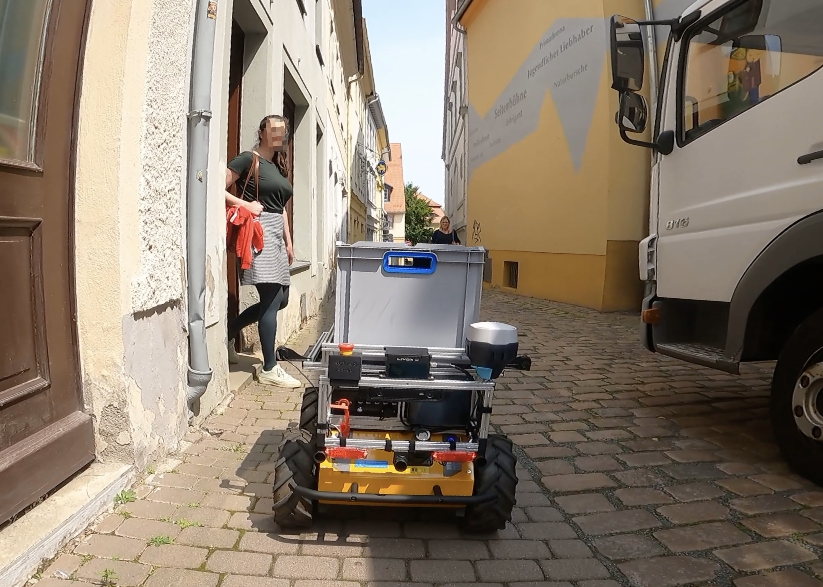}
    \caption{An exemplary conflict scenario of the Robot \textquote{Claudi} on a sidewalk.} 
    \label{fig:roboter_haustuer}
\end{figure}

\Cref{fig:roboter_haustuer} illustrates the challenge. The robot, \textquote{Claudi}, navigates a $\qty{1.2}{\meter}$ wide sidewalk restricted by a truck while a pedestrian exits a building. Although the width is sufficient for operation, it is insufficient for conflict-free passage. Such conflicts must be identified and resolved to ensure safe and socially acceptable interactions.

Two solutions are possible: adjusting the robot's path planning to avoid high-conflict areas, or adapting its behavior to the specific context to maximize social acceptance~\cite{siebert2020influence,LaucknerKM14,herzog2023perception}. Both require a deep understanding of the contextual scenarios that lead to conflict situations.
We propose a data-driven, bottom-up approach to describing interactions that enables robots to understand and resolve potential conflicts during path planning and control. This paper presents the first step towards this vision by introducing a pipeline that uses unsupervised learning to analyze robot-pedestrian interactions in urban delivery scenarios. We demonstrate its preliminary implementation with recorded robot missions and highlight the importance of incorporating contextual parameters such as junction semantics.
% In the following section, we review current research on robot-human conflicts, particularly those involving pedestrians.
% Existing studies take a human-centric, top-down approach focusing on limited parameters, such as time spent in conflict zones or proximity between humans and robots.
% Contrarily, we argue for a comprehensive set of parameters that considers the contextual scenario to capture the complexity of conflicts.
% Thus, \Cref{sec:concept} presents our proposed pipeline that takes into account static, spatial data provided by maps and dynamic, temporal data from sensors of recorded robot missions to define a data-driven approach for analyzing interactions between mobile robots and passersby.
% \Cref{sec:implementation} presents a use-case study in which we apply the pipeline to a real-world data set and discuss its results.
In the following sections, we review current research on robot-human conflicts, particularly those involving pedestrians. Existing studies often take a human-centric, top-down approach, focusing on limited parameters such as time spent in conflict zones or proximity between humans and robots. We argue for a comprehensive set of parameters that considers the contextual scenario to capture the complexity of conflicts. \Cref{sec:concept} presents our proposed pipeline that considers static spatial data from maps and dynamic temporal data from recorded robot missions. \Cref{sec:implementation} applies the pipeline to a real-world dataset and discusses the results. Finally, \Cref{sec:future_work} concludes the paper and outlines future work.
%By identifying interaction patterns known from literature in the examined data, we firstly show that we can represent interactions adequately.
%\martin{Ich bin immernoch der meinung wir haben keine interaction patterns in der Literature...}
%In a second step, we complement the interaction representation with contextual information (distance to junctions) to highlight the discriminative power of using complementary information.
Finally, \Cref{sec:future_work} concludes the paper and derives future work.

%% file: sections/01_soa.tex
\section{State of the Art}
\label{sec:soa}

The literature review by \textcite{Alverhed.2024} examines the impact of autonomous delivery robots (ADRs) on last-mile delivery, their contribution to the logistics sector, and their influence on competitive business models. Alverhed highlights four key areas affecting ADRs: operations, infrastructure, regulations, and societal acceptance. Their analysis of 113 papers reveals a lack of real-world application studies, with most applied research limited to lab or campus settings.

%\textcite{honig.2018} review 221 articles on person-following robots and identify 6 important topics which can also be applied to other robots. a) applications; b) robotic systems; c) environments; d) following strategies; e) human-robot communication; and f) evaluation methods. They point out a lack in research regarding social interaction and user needs. They also derive a set of human, robot, task and environment factors that are likely to influence people's spatial preference.

\textcite{honig.2018} review 221 articles on person-following robots and identify 6 important topics relevant across robotic applications: applications, robotic systems, environments, following strategies, human-robot communication and evaluation methods. They point out a lack in research regarding social interaction and user needs. They also derive a set of human, robot, task, and environment factors that are likely to influence people's spatial preference.

% \textcite{Arntz.2023} look at the traffic environment and the social acceptance as part of a \textquote{roboreadiness} metric. They show that infrastructure and traffic characteristics influence the social acceptance of a robot. The 4 infrastructure characteristics compose of width, alignment, crossing and elevation. The 5 traffic conditions compose of: [Type] of other road users, intensity, density, visibility and speed. They note that the limited scope and lack of consideration of difficult traffic situations may make the findings only applicable to environments similar to the Rasmus University Rotterdam campus.

\textcite{Arntz.2023} examine traffic environments and social acceptance as part of a \textquote{roboreadiness} metric, demonstrating how infrastructure and traffic characteristics affect robot acceptance. They identify four infrastructure characteristics (width, alignment, crossing, and elevation) and five traffic conditions (type of road users, intensity, density, visibility, and speed). However, they note that their study’s limited scope and exclusion of complex traffic scenarios may restrict the applicability of their findings to settings similar to the Rasmus University Rotterdam campus.

% In their study, \textcite{Gehrke.2023} deploy nine stationary cameras to observe interactions between pedestrians, cyclists, and delivery robots, focusing on potential conflicts. They use the concept of \textit{post-encroachment time} (PET) to assess the intensity of these interactions, where PET measures the time elapsed since a person has exited a potential collision zone by the time a robot enters or vice versa. A shorter PET suggests a higher risk of conflict. The researchers manually analyze 201 such interactions, detailing the nature of these encounters using \textit{conflict characteristics} (such as conflict direction, time of day and maneuver) and \textit{site characteristics} (such as exposure, presence of bicycle lane, sidewalk width, number of intersecting pathways), exploring the impact of these factors on the PET. They developed a logit model based on the three most critical factors to predict the severity level of PET. However, the study's reliance on time-based conflict evaluation and the manual method of recording interactions restrict its scope and the depth of insights into the dynamics of pedestrian-robot interactions.

In their study, \textcite{Gehrke.2023} deploy nine cameras to monitor interactions among pedestrians, cyclists, and delivery robots, using \textit{post-encroachment time} (PET) to evaluate interaction intensity. PET, which measures the time between a person leaving and a robot entering a potential collision zone or vice versa, indicates conflict risk, with shorter times suggesting higher risk. The researchers manually analyzed 201 interactions, focusing on \textit{conflict} and \textit{site characteristics} to study their impact on PET. They developed a logit model from three key factors to predict PET severity. However, the study’s reliance on time-based metrics and manual data collection limits its breadth and depth in understanding pedestrian-robot dynamics.

% The work of \textcite{Weinberg.2023} describes the interaction between people and delivery robots qualitatively on a social level. The team conducted ethnographic observations and interviews to capture pedestrians' perceptions of the delivery robots during a pilot project. The authors systematically recorded conflict situations in which the robots distracted and obstructed various sidewalk users (including children and dogs), as well as how people helped or hindered the robots in return. The recording of interactions is purely manual using pen and paper. The interactions are analyzed individually and there was no comprehensive summary of the influencing parameters.

The study by \textcite{Weinberg.2023} qualitatively explores social interactions between people and delivery robots. The researchers used ethnographic observations and interviews during a pilot project to understand pedestrians' perceptions. They manually recorded instances where robots caused distractions or obstructions for sidewalk users, including children and dogs, and noted how people either assisted or impeded the robots. These interactions were analyzed individually, without a comprehensive summary of influencing factors.

% A procedure for characterizing parameters that promote conflicts is described in the paper by \textcite{Mayerhofer.2020}. The paper classifies static and dynamic environmental parameters (weather, light, other road users, etc.) and links them to the topology of the routes. However, it derives an approach based on literature studies, brainstorming, and expert discussions and lacks a quantitative statistics.

% \textcite{JeredVroon.2020} conducts observations and discusses when an interaction turns into a conflict. However only 5 observations are listed.

% Other studies observe the lateral and medial distance of a robot \cite{ShiannevanMierlo.16.06.2021, pacchierotti.2006, neggers.2022.size, neggers.2018} and the impact on human acceptance or comfort. Studies show that the speed of the robot influences the passing distance. \cite {zhang.2019, neggers.2022.speed}
\textcite{Mayerhofer.2020} describes a method for identifying conflict-promoting parameters, classifying them as static or dynamic (weather, lighting, road users, etc.), and linking them to route topology. The approach is developed through literature review, brainstorming, and expert discussions but lacks quantitative analysis.

\textcite{JeredVroon.2020} discusses how interactions escalate into conflicts, but only five observations are documented.

Other research focuses on the lateral and medial distances maintained by robots \cite{ShiannevanMierlo.16.06.2021, pacchierotti.2006, neggers.2022.size, neggers.2018} and their effect on human comfort or acceptance. Other studies indicate that a robot's speed affects the distance people keep while passing \cite{zhang.2019, neggers.2022.speed}.
%, which does not represent the complex behavior in Human-Robot interactions in a real-world scenario.

%\textcite{hiroi.2019} shows that the size of the robot itself has an influence on the anxiety people feel when interacting. Similarly \cite{jost.2021} shows the distance people keep to the robot increases with size and is reduced by a friendly expression of the robot.

%% Summary of SoA
% \todo[inline]{Add details from the section - which metrics are used to assess the quality of interactions? What types of interactions are possible? What identifies good/bad interactions?}
To summarize, the current studies focus primarily on the outside examination of the robot. 
No research was identified which used the sensors of the robot to investigate the situation. 
Thus, it is not yet known what information about the environment can be gathered from the robot's point of view.
Similarly, there is no consensus on a metric for assessing the quality of an interaction. 
%Similarly, there is no consensus on \replaced{a metric for assessing the quality of an interaction}{what constitutes a \textquote{good} or \textquote{bad} interaction for the robot}. 
In general, the types of interactions that can take place in a real-world scenario are not well understood.
% keiner machts aus Roboter Sicht! Alle gucken immer nur von draußen drauf.
% Der Mensch steht im Fokus, das ist Ethisch auch ok, aber für neue erkentnisse muss man es auch mal aus Roboteraugen betrachten.
% Das der umgebende Kontext relevant ist, zeigt sich durch die wiederholte Erwähnung in den verschiedenen Papern, aber welche Parameter welchen Einfluss haben ist dabei noch unklar.
% Was gut oder schlechte interaktionen sind und was nicht wird überwiegend aus Menschensicht beeurteilt, nur wenige Paper betrachten auch z.B. den Zeitverlust des Roboters als thematik.
% Welche Interaktionen es tatsächlich in der realen Welt gibt ist noch nicht abschließend geklärt. 

% To summarize the state of the art, we note that there are no universal legal guidelines for mobile robot interactions with pedestrians on shared paths -- regulations vary depending on the location.  
% Current methods for evaluating potential conflict situations consider either time or distance alone, favoring one perspective over the other and thereby failing to capture the full spectrum of relevant parameters.
% Existing frameworks for defining interactions prioritize human-centric, top-down approaches, neglecting valuable insights that could be gained from data-driven, bottom-up methods that classify interactions without human bias.

%% file: sections/02_concept.tex
\section{Concept}
\label{sec:concept}

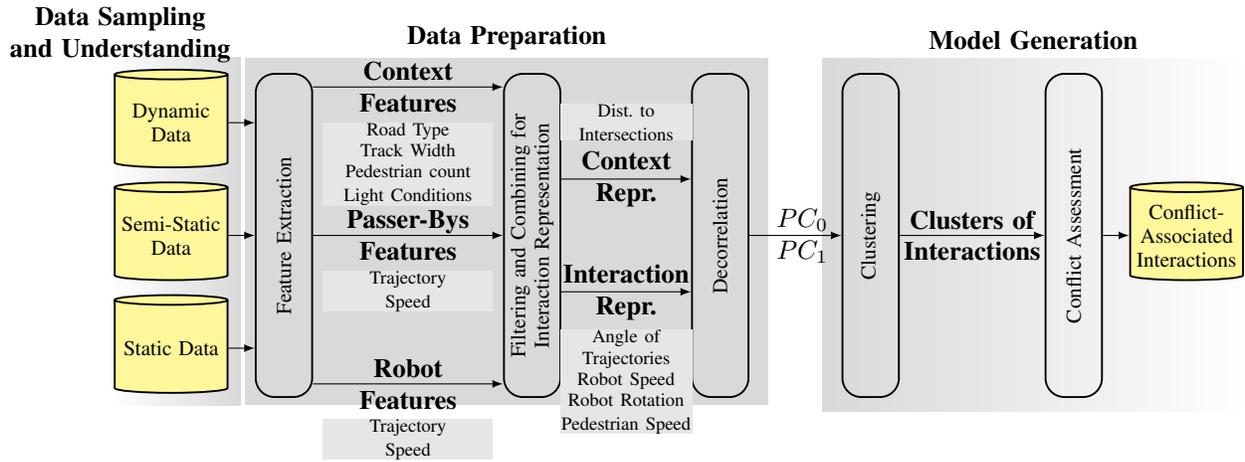
\begin{figure*}[t]
  \centering
  \input{figures/tikz/pipeline_dfd.tex}
  \caption{Pipeline for data-driven, bottom-up analysis of human-robot interactions in urban delivery scenarios.}   
  \label{fig:pipeline}
\end{figure*}

Our vision is to analyze interaction patterns encountered during the robot's operation to identify potential for conflicts. 
These can then be used for mission planing and optimizing behavior during its execution.
%Our vision is to cluster the robot's area of operation according to the interaction patterns expected there. In this way, the individual conflict potential becomes part of the planning process before the mission or can be used during execution to optimize the robot's behavior.
To implement such awareness for interactions, the first step is to define parameters required to represent interaction scenarios. 
For that, we explore which static, semi-static, and dynamic information must be captured by these parameters to sufficiently represent interactions by defining a pipeline using unsupervised learning.

The proposed pipeline is schematically displayed in \Cref{fig:pipeline}. 
Note that it shows the overall concept while the current implementation discussed in \Cref{sec:implementation} covers only core parameters.
%The diagram shows the overall concept to be implemented in the medium term; the current implementation only covers the core parameters of the potentially suitable features and associated extractors as proof of concept.
We detail each phase in the next subsections.
%Firstly, data sampling and understanding describes the categories of data sources considered as potential inputs for our pipeline.
%In \Cref{subsec:sampling}, we describe the data preparation used to extract potential features capturing the robot's and pedestrian's movements as well as the contextual situation. 
%Building up on these features, we propose filtering and combining central features in \Cref{subsec:feature} to represent interactions before de-correlating the data for preparing clustering in the phase of Model Generation, \Cmpr{subsec:modeling}.

\subsection{Data Sampling and Understanding}
\label{subsec:sampling}

The goal of the pipeline is to explore and identify relevant parameters to describe interactions between mobile robots and pedestrians.
% As described in \Cref{sec:soa}, central parameters considered in literature are the distance between participating entities~\cite{ShiannevanMierlo.16.06.2021, pacchierotti.2006, neggers.2022.size, neggers.2022.speed, neggers.2018}, the time between entities entering spatial zones of conflicting trajectories~\cite{Gehrke.2023}, and trajectories of the entities interacting with each other~\cite{ShiannevanMierlo.16.06.2021, neggers.2022.speed}.
% These parameters and their corresponding studies, however, focused on human-centric, top-down approaches and partially stem from articifial/laboratory settings. 
% In contrast, our goal is to enable robots to be aware of interactions, which requires a data-drive, bottom-up approach. 
% Therefore, data for analyzing interactions should include contextual information as well. 
For that, we distinguish three types of data as inputs to our pipeline.

\paragraph{Static}
%Following our hypothesis, contextual parameters are needed to describe interactions. 
Static data is assumed to not change over time. 
In addition to parameters describing the interaction itself, static parameters such as the spatial layout in which the interaction takes place are central.  
This includes geometric information (e.g. positions and distances of landmarks, points of interest, and infrastructure) as well as abstracted semantic information (e.g. the type of road or whether both entities are at a junction). 

\paragraph{Semi-Static}
Semi-static data is assumed to be static for the time of the interaction but may still vary over time. 
Examples are records of weather or lighting conditions, which are changing with reduced frequency.
The reason for explicitly considering this source of data is their potential to identify temporal dependencies between the parameters and the characteristics of interactions. 

\paragraph{Dynamic}
Dynamic data yields information that describes the development of a conflict situation over time. 
Such as the trajectories of the robot and pedestrians as well as contextual parameters that change during the course of a conflict situation. 
Potential sources are the perception system of the robot itself as well as sensors of attributed environments.

For each type of data sources, multiple providers can be integrate to enhance the overall dataset. For instance, multiple map data services might be used to obtain static data and overcome lack of completeness in individual providers~\cite{Plank.2022}. Similarly, the integration of dynamic data from diverse sources, such as the robot's perception system and sensors of the infrastructure (e.g. traffic cameras), can further enrich the understanding. The subsequent \textit{Data Preparation} phase makes it possible to exploit the temporal characteristics of different sources, such as applying complementary fusion.
In this way, the strict separation between the initial types of data sources can be dissolved.
%For each type of data sources, multiple actual sources might be considered to complement each other.
% For instance, multiple map data services might be used to obtain static data and overcome lack of completeness in individual providers~\cite{Plank.2022}.
% Moreover, this lack can be addressed by dynamic data stemming from multiple sources (robot's perception system, sensors of attributed infrastructure, $\ldots$) again.
% The following \textit{Data Preparation} phase allows leveraging the temporal characteristics of different sources by applying, for instance, complementary fusion.
% Thereby the strict separation between initial types of data sources can be dissolved.

\subsection{Data Preparation}
\label{subsec:feature}

%\todo[inline]{Which parameters/features are listed in the state of the art for representing/characterizing conflict situations?}

For exploring the space of interactions while focusing on conflict situations, we need to
\begin{enumerate*}
    \item extract relevant features
    \item filter and combine them to represent interactions
    \item normalize and de-correlate the data to prepare clustering.
\end{enumerate*}
Each of these steps are part of the data preparation and is detailed in this subsection.

%\todo[inline]{Find a better way to integrate the structure in the text -- paragraphs and italic texts are not working and do not visually structure the text nicely. Its confusing to the reader}
\subsubsection{Feature Extraction}
% \paragraph{Feature Extraction}
The first step is to process the available data to extract relevant features.
To structure them, we consider three categories of features, \Cmpr{fig:pipeline}.

\begin{enumerate}[label=(\alph*)]
    \item \textit{Context Features} include the environmental setting and infrastructure relevant to where the interaction takes place.
The spatial layout and information about the surrounding infrastructure may be obtained from static data while weather conditions or temporary obstacles (e.g. parked cars or trash bins) may be available as semi-static data. 
    \item \textit{Passersby Features} describe the pedestrian or cyclist involved in the interaction.
Central to this category are the trajectory, speed and orientation of the individual, possibly supplemented by demographic and behavioral data (e.g. gender, age, emotional expressions) if available.
This information may be available as dynamic or semi-static data depending on the acquisition process and level of abstraction.
    \item \textit{Robot Features} detail the characteristics and behavior of the robotic system during the interaction.
Key features include the robot's trajectory, speed and orientation, commonly extracted from dynamic data sources, such as the robot's own perception system.
\end{enumerate}

With these features, the feature extraction combines information from different sources and standardizes access.
%The data aggregation level brings together information from different sources and standardizes access.

\subsubsection{Representation of Interactions and Contexts}

This step aims at transforming extracted features to obtain abstract representations of context and interactions.
To represent interactions, features related to both, the robot and the pedestrians, are synchronized. This may involve estimating intermediate positions and fitting mathematical models to represent trajectories.
For contextual representation, calculations could include measuring sidewalk widths, determining distances to house entrances, and counting pedestrians in the vicinity of the robot.
In addition, statistical values such as maximum, minimum, mean and standard deviation can be aggregated to illustrate the variability of features.

The exploratory nature of the pipeline enables the dynamic definition of these features with varying spatial and temporal boundaries, allowing us to investigate their impact.
This approach supports the analysis and helps to address specific questions, such as determining the significance of including junctions as a contextual feature based on their proximity, defined by a threshold distance of $\qty[parse-numbers=false]{x}{\meter}$.

\subsubsection{Normalization and De-correlation}

Combining the representation of context and interaction, each interaction scenario is described by an $10$-dimensional vector. 
In order to prepare clustering the data to explore patterns and identify categories of interactions, the data needs to be normalized and de-correlated.

Firstly, we apply Z-Score normalization $x' = (x-\mu)/\sigma$ of the data set to transform the $10$D vectors to have a standard deviation of $1$ in each dimension.
This normalization ensures that each dimension contributes equally to the distance metric used in clustering, thus providing equal weighting to all parameters. 
Prior to clustering, we employ Principal Component Analysis (PCA)~\cite{hastie2009elements} to reduce the dataset's dimensions.
Again, noting the work-in-progress characteristics of the current pipeline's state, we reduce the number of dimensions to 2D to enable simplified visualizations.

\subsection{Model Generation}
\label{subsec:modeling}

Having the data normalized and reduced to 2D, model generation can be applied to explore interaction patterns. 
For that, we cluster the data using K-means clustering~\cite{hastie2009elements} and assess to which degree they contain interactions of conflicts with pedestrians using the distance between both entities. 
This shall allow us to differentiate between clusters representing critical interactions indicating conflict situations that potentially reduce social acceptance and interactions that represent positive encounters with acceptable behavior.

%% file: figures/tikz/pipeline_dfd.tex
% define command to draw a data base and its label together
% \drawdatabase{<name>}{<position>}{<label>}
\newcommand{\drawdatabase}[3]{
  \node[database] (#1) at (#2) {};
  \node at (#1) [text width=1.35cm, align=center] {#3};
}

\newcommand{\xshift}{2cm} % definition of xshift
\newcommand{\yshift}{1.5cm} % definition of yshift

% define background layer for the diagram
\pgfdeclarelayer{background}
\pgfdeclarelayer{foreground}
\pgfsetlayers{background,main,foreground}   %% some additional layers for demo

\usetikzlibrary{fadings}

\begin{tikzpicture}[
    node distance=2.2cm,
    database/.style={
      cylinder,
      minimum width = 1.5cm,
      minimum height = 1.3cm,
      cylinder uses custom fill,
      cylinder body fill=yellow!50,
      cylinder end fill=yellow!50,
      shape border rotate=90,
      aspect=0.55,
      draw
    },
    process/.style={
      rectangle,
      rounded corners=0.25cm,
      minimum width = 1.5cm,
      minimum height = 0.75cm,
      draw, inner sep=0,
    },
    every node/.style={
        font=\footnotesize,
        line width=0.25mm
    },
  ]
    \drawdatabase{static_data}{$(0,0)$}{Static Data}
    \drawdatabase{semi_static_data}{$(static_data) + (0, \yshift)$}{Semi-Static Data}
    \drawdatabase{dynamic_data}{$(semi_static_data) + (0, \yshift)$}{Dynamic Data}

    % draw rectangle with rounded corners that spans the height of all three databases
    \node[process, minimum height=0.75cm, text width=2*\yshift+1.3cm, align=center, rotate=90] (feature_extraction) at ($(semi_static_data)+(0.75*\xshift,0)$) {Feature Extraction};

    % Nodes for Interaction representation using features, decorrelation using PCA and clustering
    \node[process, minimum height=0.75cm, text width=2*\yshift+1.3cm, align=center, rotate=90] (interaction_representation) at ($(feature_extraction) + (1.65*\xshift,0)$) {Filtering and Combining for Interaction Representation};
    \node[process, minimum height=0.75cm, text width=2*\yshift+1.3cm, align=center, rotate=90] (decorrelation) at ($(interaction_representation) + (1.25*\xshift,0)$) {Decorrelation};
    \node[process, minimum height=0.75cm, text width=2*\yshift+1.3cm, align=center, rotate=90] (clustering) at ($(decorrelation) + (1.0*\xshift,0)$) {Clustering};
    \node[process, minimum height=0.75cm, text width=2*\yshift+1.3cm, align=center, rotate=90] (classification) at ($(clustering) + (1.35*\xshift,0)$) {Conflict Assessment};

    \drawdatabase{conflict_interactions}{$(classification)+(0.75*\xshift,0)$}{Conflict-Associated Interactions}

    % edges -- to feature extraction
    \path let \p1=(static_data), \p2=(feature_extraction.north) in coordinate (fe_input_static) at (\x2,\y1); 
    \draw[-latex] (static_data) -- (fe_input_static);

    \path let \p1=(semi_static_data), \p2=(feature_extraction.north) in coordinate (fe_input_semi) at (\x2,\y1); 
    \draw[-latex] (semi_static_data) -- (fe_input_semi);

    \path let \p1=(dynamic_data), \p2=(feature_extraction.north) in coordinate (fe_input_dynamic) at (\x2,\y1); 
    \draw[-latex] (dynamic_data) -- (fe_input_dynamic);

    % edges -- from feature extraction to interaction representation
    \draw[-latex] ($(feature_extraction.south east) + (0,-0.125*\yshift)$) -- ($(interaction_representation.north east) + (0,-0.125*\yshift)$) node[midway, text width=3cm, align=center, font=\bfseries] (context_parameters) {Context\\Features};
    \node[text width=2.25cm, align=center, anchor=north, fill=black!10, inner sep=0] at ($(context_parameters.south) + (0,0*\yshift)$) {\scriptsize{Road Type\\Track Width\\Pedestrian count\\Light Conditions}};

    \draw[-latex] ($(feature_extraction.south) + (0,-0*\yshift)$) -- ($(interaction_representation.north) + (0,0*\yshift)$) node[midway, text width=3cm, align=center, font=\bfseries] (passers_parameters) {Passer-Bys\\Features};
    \node[text width=2.25cm, align=center, anchor=north, fill=black!10, inner sep=0] at ($(passers_parameters.south) + (0,-0*\yshift)$) {\scriptsize{Trajectory\\Speed}};

    \draw[-latex] ($(feature_extraction.south west) + (0,0.125*\yshift)$) -- ($(interaction_representation.north west) + (0,0.125*\yshift)$) node[midway, text width=3cm, align=center, font=\bfseries] (robot_parameters) {Robot\\Features};
    \node[text width=2.25cm, align=center, anchor=north, fill=black!10, inner sep=0] at ($(robot_parameters.south) + (0,-0*\yshift)$) {\scriptsize{Trajectory\\Speed}};

    % edges -- from interaction representation to decorrelation
    \draw[-latex] ($(interaction_representation.south)+(0,0.5*\yshift)$) -- ($(decorrelation.north)+(0,0.5*\yshift)$) node[midway, text width=3cm, align=center, font=\bfseries] (context_data) {Context\\Repr.};
    \node[text width=1.75cm, align=center, anchor=south, fill=black!10, inner sep=0] at ($(context_data.north) + (0,0*\yshift)$) {\scriptsize{Dist. to Intersections}};
    \draw[-latex] ($(interaction_representation.south)+(0,-0.5*\yshift)$) -- ($(decorrelation.north)+(0,-0.5*\yshift)$) node[midway, text width=3cm, align=center, font=\bfseries] (interaction_data) {Interaction\\Repr.};
    \node[text width=1.75cm, align=center, anchor=north, fill=black!10, inner sep=0] at ($(interaction_data.south) + (0,-0*\yshift)$) {\scriptsize{Angle of Trajectories\\Robot Speed\\Robot Rotation\\Pedestrian Speed}};

    % edges -- from decorrelation to clustering
    \draw[-latex] ($(decorrelation.south)$) -- ($(clustering.north)$) node[midway, text width=0.5cm, align=center, font=\bfseries] (pcs) {$PC_0$\\$PC_1$};

    % edges -- from clustering to classification
    \draw[-latex] ($(clustering.south)$) -- ($(classification.north)$) node[midway, text width=2cm, align=center, font=\bfseries] (clusters) {Clusters of Interactions};

    % edges -- from classification to output
    \draw[-latex] ($(classification.south)$) -- ($(conflict_interactions.west)$);

    % draw shaded areas on background layer to visualize steps of the pipeline
    \begin{pgfonlayer}{background}
        % draw shaded area for data sampling and understanding
        \filldraw[fill=black!15,draw=none, path fading = west] ($(static_data.south west)+(-0.125*\xshift, -0.125*\yshift)$) rectangle ($(dynamic_data.north east)+(0.125*\xshift, 0.125*\yshift)$);
        % add label
        \node[anchor=south, text width=3cm, align=center, font=\bfseries] at ($(dynamic_data.north west)$) {Data Sampling and Understanding};

        % draw shaded area for data preparation
        \coordinate (south_west) at ($(feature_extraction.north west)+(-0.06125*\xshift, -0.06125*\yshift)$);
        \coordinate (north_east) at ($(decorrelation.south east)+(0.125*\xshift, 0.125*\yshift)$);
        \draw[fill=black!15,draw=none] (south_west) rectangle (north_east);
        % add label
        \path let \p1=(south_west), \p2=(north_east) in coordinate (label_position) at ($(\x1, \y2)!0.5!(north_east)$);
        \node[anchor=south, text width=3cm, align=center, font=\bfseries] at (label_position) {Data Preparation};

        % draw shaded area for data preparation
        \coordinate (south_west_modelling) at ($(clustering.north west)+(-0.125*\xshift, -0.125*\yshift)$);
        \path let \p1=($(classification.south east)+(0.125*\xshift, 0.125*\yshift)$), \p2=(conflict_interactions.east) in coordinate (north_east_modelling) at ($(\x2, \y1)$);
        \filldraw[fill=black!15,draw=none, path fading = east] (south_west_modelling) rectangle (north_east_modelling);
        % add label
        \path let \p1=(south_west_modelling), \p2=(north_east_modelling) in coordinate (label_position_modelling) at ($(\x1, \y2)!0.5!(north_east_modelling)$);
        \node[anchor=south, text width=3cm, align=center, font=\bfseries] at (label_position_modelling) {Model Generation};

    \end{pgfonlayer}

  \end{tikzpicture}

%% file: sections/03_implementation.tex
\section{Use Case: The RoboTraces Dataset}
\label{sec:implementation}

The last section introduced our pipeline for exploring interaction patterns of robots and pedestrians in urban environments conceptually. 
In this section, we present a use-case study in which we apply the pipeline to our \textit{RoboTraces} dataset \cite{Zug.91220239152023}.
We follow the same structure as before.
We firstly discuss the datasets used as inputs to the pipeline before we describe the calculations for data preparation in \Cref{subsec:use_case_preparation}.
The clustering after normalization and de-correlation is discussed in \Cref{subsec:use_case_model_generation}.
Finally, \Cref{subsec:results} discusses the results of applying the pipeline.

%%%%%%%%%%%%%%%%%%%%%%%%%%%
%%% Data Sampling and Understanding
%%%%%%%%%%%%%%%%%%%%%%%%%%%
\subsection{Data Sampling and Understanding}
\label{subsec:use_case_sampling}

The pipeline considers static, semi-static, and dynamic data as inputs.
While Open Street Map (OSM)\footnote{\url{https://www.openstreetmap.org}} is considered for static data, data sets recorded during operating our robot \textit{Claudi} \Cmpr{fig:roboter_haustuer} in the city center of Freiberg are considered dynamic data. 
Both data sources are discussed in the following.
Semi-static data is not considered in this work-in-progress study.

\paragraph{Claudi}

%For aggregating dynamic data, we defined an track for our robot \textit{Claudi} to operate on in the city center of Freiberg, Germany. 
Claudi carries a cargo box to resemble a delivery robot and measures $\qty{100}{\cm}$ in length, $\qty{67}{\cm}$ in width, and $\qty{100}{\cm}$ in height.
It is equipped with four stereo cameras (ZED 2i\footnote{\url{https://store.stereolabs.com/en-de/products/zed-2i}, last accessed \formatdate{9}{4}{2024}}) allowing for a $\qty{360}{\degree}$ field-of-view around the robot.
To locate Claudi, it is equipped with a multi-band Real-Time Kinematic (RTK) Global Navigation Satellite System (GNSS) (Emlid Reach RS2\footnote{\url{https://docs.emlid.com/reachrs2/specs/}, last accessed \formatdate{9}{4}{2024}}) that provides positioning with a precision of $H: \leq \qty{7}{\milli\meter}+1~ppm, V: \leq \qty{14}{\milli\meter}+1~ppm$.
The track Claudi was manually operated on for data acquisition (partially shown in \Cref{fig:track}) has a length of $\qty{2190}{\meter}$ for which Claudi needed an average time of $\qty{62}{\minute}$ to finish.
The track covers different configurations of sidewalks and bicycle lanes (pedestrian zones, open places with obstacles, narrow cobblestone paths in the medieval area of the city).
We recorded 24 missions as described in \cite{Zug.91220239152023}. 
Out of these, 5 could be used for this study.
All records are stored as ROS2~\cite{ros2} bag files in a database system\footnote{An exemplary dataset will be available on \url{https://mobilithek.info/offers/705118540886962176} }. 
The recorded missions were scheduled to cover not only all calendrical seasons, but also to capture interactions with pedestrians during day- and night-time.
This enriches the number of contextual parameters captured and allows (although not used in this work-in-progress study) to extract semi-static data as well.

\paragraph{Open Street Map}

To complement the dynamic data provided by the recorded robot missions, we aggregate static data from the open data platform OSM -- most importantly semantic information about the spatial environment interactions take place in.
Existing frameworks provide a rich environment for accessing the database and simplify automated extraction of information, which supports the feature extraction phase.
In this work-in-progress study, we leverage OSM to analyze whether proximity to junctions can distinguish between types of interactions involving the entities.

%It should be noted at this point that the strict separation described above—static data from a map database and dynamic data from the robot—is dissolved for more detailed investigations with additional context features. Due to the lack of completeness of map data services regarding the relevant static information, the robot must also aggregate these parameters~\cite{Plank.2022}.
%\georg{I moved this discussion to the description of data sources in the concept part. I think that the lack of completness is actually to motivation for us to use different data sources in the first place...}

%%%%%%%%%%%%%%%%%%%%%%%%%%%
%%% Data Preparation
%%%%%%%%%%%%%%%%%%%%%%%%%%%
\subsection{Data Preparation}
\label{subsec:use_case_preparation}

The data preparation phase employs feature extraction on the static and dynamic data sources. passerby, robot, and contextual features are first extracted and subsequently combined to generate context and interaction representations. This subsection details the implemented calculations of features and their fusion for representation generation.
%Having the static and dynamic data sources in place, the data preparation phase applies feature extraction to firstly generate passersby, robot, and contextual features before these are combined to derive context and interaction representations.
%In this subsection, we detail the calculation of the individual features and how they are combined to generate the context and interaction representations.

\paragraph{Feature Extraction}

%The pipeline's central objective is to explore interaction patterns. 
%Thus, we firstly need to identify the parts of the recorded robot missions constituting robot-pedestrian interaction scenarios.
To identified relevant segments within the recorded missions, we firstly leverage the object detection framework of the ZED 2i cameras. It detects pedestrians and assigns unique IDs for tracking them within the camera's field-of-view. These IDs, along with timestamps of the first and last detection, are used to filter the data and associate sensory information with corresponding interaction scenarios. This process identified 987 interaction scenarios within the recorded dataset.

%For representing robot-pedestrian interaction, we exploit the object detection framework provided for the ZED 2i cameras.
% It detects persons and assigns unique IDs such that they can be tracked within the field-of-view of each camera. 
% We use the unique ID and its first and last detection time to filter the recorded robot missions and associate the sensory data to the corresponding interaction scenario.
% With that, we identified 987 interaction scenarios in the recorded data.

Based on the unique ID of each detected person, we extract the \textit{Passersby Features}.
We use the GNSS data of the robot to obtain a global positioning and apply coordinate transformation using the distance measurement of the detected person to generate the trajectory of the pedestrian relative to the robot.
This additionally allows us to estimate the pedestrian's velocity, which we average over the duration of the interaction scenario.
Thus, for each interaction scenario we obtain the pedestrian's trajectory and its mean velocity.

For extracting the \textit{Robot Features} we filter the GNSS and odometry data. 
We calculate the robot's trajectory during each interaction scenario and derive its mean translational and mean rotational velocities. 

In this study, \textit{Context Features} are restricted on the proximity of the robot and pedestrian to the nearest junction. We utilize the \textit{OSMnx} package \cite{BOEING2017126} to extract junctions from a simplified graph of the local road network obtained via the OSM API. This graph retains only nodes representing junctions. By parsing the trajectories of both, the pedestrian and the robot, in each interaction scenario, we can calculate their distance to the nearest junction.

\paragraph{Context and Interaction Representation}

\begin{figure}[h]
        \centering
        % trim={<left> <lower> <right> <upper>}
        %\includegraphics[trim={2cm 0.6cm 5cm 17cm},clip,width=0.5\textwidth]{figures/map_freiberg.jpg}
        \includegraphics[width=0.45\textwidth]{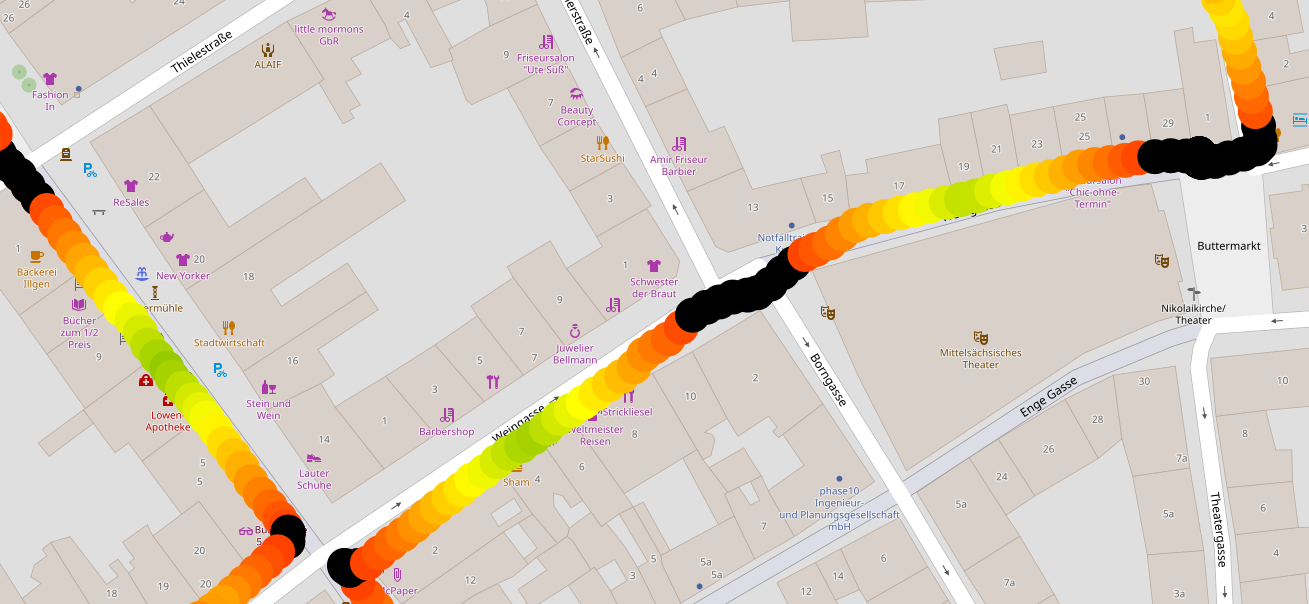}
        \caption{Part of the robot's trajectory in Freiberg, it covers different parts of the city (city center, low-traffic side streets, park). For each point at the trajectory, a distance information to the next junction node is calculated ranging from green at $\qty{65}{\meter}$ to red at $\qty{0}{\meter}$ distance. Positions closer than $\qty{8}{\meter}$ to a junction are highlighted by black color.}
        \label{fig:track}
\end{figure}

The feature extraction provides robot, pedestrian, and context features for each interaction scenario.
Those are further processed and combined to obtain context and interaction representations for each of the interaction scenarios. 

For representing the context, we focus on the distance of the robot and pedestrian to the next junction.
We consider the minimal, maximal, and mean distance to explore the importance of different values.  \Cref{fig:track} illustrates these distances along the robot's track.
This results in a $6$-dimensional vector describing the context representation. 

For representing interactions, we combine the robot's and the passerby's features. 
We concatenate the $1$D values of the robot's translational and rotational velocities with the estimated speed of the pedestrian. 
For these values, we consider the mean values over the course of the interaction scenario.
The final value to describe the interaction is the angle $\alpha$ between the directions of movements of the robot and the pedestrian~\Cmpr{fig:trajectory_examples}.
Applying linear regression to the points comprising both trajectories, we obtain two vectors (one for the robot, one for the pedestrian) describing their movements in a 2D space.
The angle $\alpha$ between the movement's directions is therefore the angle between those two vectors.
The underlying linear regression, however, represents a strong assumption and limits the presented approach as it is valid only for a subset of interactions.
%\Cref{fig:trajectory_negative_examples} exemplary visualizes interactions for which the trajectories can not be approximated as lines. 
Using the Pearson correlation coefficient obtained during linear regression, we filter all interaction scenarios and remove those not fitting the assumption of linearity, that is, for which $|r_p| \leq 0.5$. 
This removes 201 interaction scenarios, leaving a total of 786.
Alternative representations need to be explored in future work.

\begin{figure}
    \centering
    \includegraphics[width=0.5\textwidth]{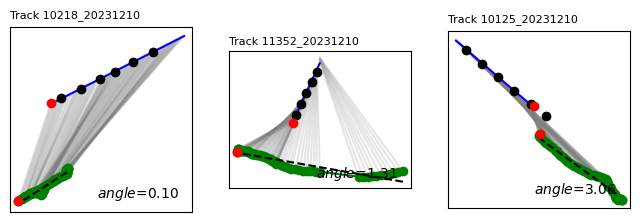}
    \caption{Movements of pedestrians and the robot in different spatial constellations: a parallel movement, a lateral encounter and a head-on confrontation. Black dots represent the robot's GNSS positions. Human movement is shown by green dots. In both cases the red dot denoting the latest human position. Gray lines illustrate the connection between the robot's observation points and the human's positions. $\alpha$ indicated the angle between the two movements. Ranging from 0 indicating parallel movement to $\pi$ head-on movement.}
    \label{fig:trajectory_examples}
\end{figure}

\begin{figure} [t]
    \centering
    \includegraphics[width=0.5\textwidth]{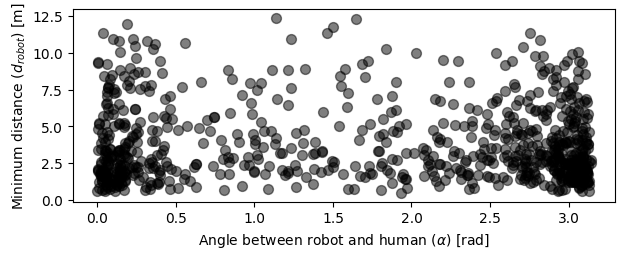}
    \caption{Scatter plot of the minimum distance between robot and pedestrian and the orientation of the directions of movement ($\qty{0}{\radian}\leq\alpha\leq\qty[parse-numbers=false]{\pi}{\radian}$)}
    \label{fig:distribution}
\end{figure}
To preliminary evaluate the discriminatory power of the angle $\alpha$, we use the minimum distance between the robot and the pedestrian ($d_{robot}$) and display it depending on the angle $\alpha$ in a scatter plot \Cmpr{fig:distribution}.
%Additionally, we mark interaction scenarios where pedestrians move slower than $\qty{0.4}{\meter\per\second}$.
The resulting plot is dominated by two poles representing parallel movements ($\alpha<\qty{0.5}{\radian}$) and head-on situations ($\alpha>\qty{2.7}{\radian}$).
%\todo[inline]{Can we have the mean/median values of the distance between robot and human for both poles? 
%$d_{robot}<0.3$ median = 2.840, mean = 3.7598
%$d_{robot}>2.8$ median = 2.862, mean = 3.6297
%}
For those, the average distance between the robot and pedestrian is reduced compared to scenarios in which the entities are approaching each other in a lateral movement ($\qty{0.5}{\radian}\leq\alpha\leq\qty{2.7}{\radian}$).
These scenarios occur scarcely compared to the poles and show an uniform distribution -- both considering the angle $\alpha$ and the distance between pedestrian and robot.
Nevertheless, the visible poles provide an indication for the discriminatory power of the angle $\alpha$, which constitutes the last component of the interaction representation.
%Thus, each interaction is represented by a $4$D vector.

Concluding the interaction representation, we concatenate both vectors, the context and interaction representations, to obtain a $10$-dimensional vector for each of the 786 interaction scenarios.
After normalization and de-correlation as described in \Cref{subsec:feature}, these scenarios are represented by their first and second principal components, $PC_0$ and $PC_1$.

% This data set is then normalized and de-correlated as described in \Cref{subsec:feature}.
% This results in 786 interaction scenarios represented by their first and second principal component $PC_0$ and $PC_1$. The data set is passed on to the next phase, the model generation.

%%%%%%%%%%%%%%%%%%%%%%%%%%%
%%% Model Generation
%%%%%%%%%%%%%%%%%%%%%%%%%%%
\subsection{Model Generation}
\label{subsec:use_case_model_generation}

\begin{figure}
    \centering
    \includegraphics[width=0.5\textwidth]{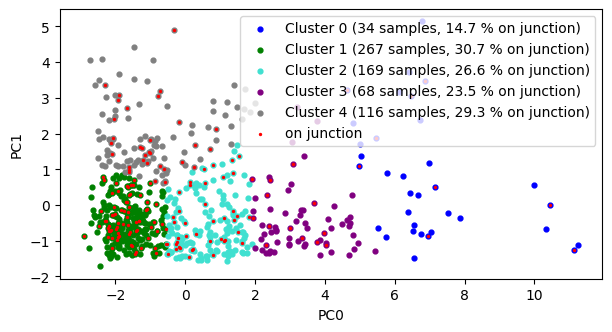}
    \caption{Clustering interaction scenarios after dimensionality reduction using PCA with K-means ($k=4$). The red dots indicate interaction scenarios at junctions.}
    \label{fig:kmeas_clustering}
\end{figure}

The phase of model generation leverages the prepared data and applies unsupervised learning to explore the patterns of interactions between the robot and pedestrians.
In this use-case study, we apply K-means clustering with the euclidean norm as a distance metric~\cite{hastie2009elements}.
%\todo[inline]{Which number of clusters did we evaluate? \textit{2-9 habe ich ausprobiert und immer die maximale Summe der Abweichung der entstandenen Verteilungen aufgestellt. Bei 4 gab es ein maximum.}}
%\todo[inline]{Which metric for comparing the distributions did we use?}
The minimum distance to passersby was again used as an indicator of the quality of the results. 
We evaluated values for $k \in [3,9]$ and assessed the quality of clustering using the Kolmogorov–Smirnov (KS) statistic ($D_n$) and Wasserstein distance $W_1$ on the resulting distributions of $d_{robot}$. 
The maximum distance $max(D_n)=0.23$, that is, the maximal discriminatory power is achieved when clustering with $k=5$ clusters.

\subsection{Results and Discussion}
\label{subsec:results}

Based on the pipeline's results, this section compares the proposed data-driven, bottom-up approach to segmenting interaction scenarios with an human-centric, top-down alternative.

For that, we revisit the contextual factor of junction/non-junction scenarios.
An interaction scenario is assigned a junction scenario if the robot and involved pedestrian are closer than $\qty{8}{m}$ to the nearest junction's geometric center.

\Cref{fig:kmeas_clustering} shows the 5 clusters and marks the scenarios that took place at a junction according to that classification rule. 
We can make two observations.
Firstly, the sample count per cluster differ significantly. 
Secondly, the junction scenarios are distributed across all clusters, but vary in their ratio to non-junction scenarios. 
Thus, the segmentation we are arriving at when using a bottom-up, data-driven approach differs significantly from the top-down, human-centric segmentation.
In other words, the robot focuses on other aspects of interactions than the human-centric definition suggests.

%The allocation of the scenarios in the clusters according to the top-down and bottom-up approach differs significantly.
%\sebastian{Das unerfüttere ich noch mit Zahlen in der Grafik.}
%\georg{Leider verstehe ich hier die Aussage der Diskussion nicht? Was ist die Konsequenz?...das die Frage dann im Text auftaucht, hilft mir leider nicht!}

\begin{figure}
    \subfigure[Clusters identified through K-means clustering.]
    {
    \centering
        \includegraphics[width=0.22\textwidth]{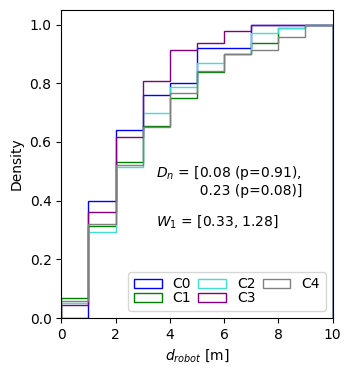}
        \label{fig:ecdf_clustering}
    }
    \subfigure[Junction and non-junctions scenarios]
    {
    \centering
        \includegraphics[width=0.22\textwidth]{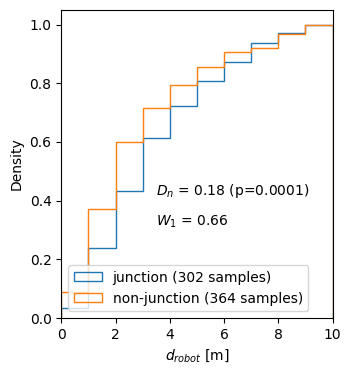}
        \label{fig:manual_classification}
    }
        \caption{Empirical cumulative distribution functions (ECDFs) of minimum distances between robots and pedestrians}
        \label{fig:three graphs}
\end{figure}

We can examine the results further by analyzing the minimum distance between passersby and the robot within the classes obtained through the top-down and bottom-up approaches.
\Cref{fig:ecdf_clustering} depicts the empirical cumulative distribution functions (ECDFs) for the bottom-up approach. 
Taking this perspective does not show significant differences between the clusters, that is, the human-centric perspective does not allow to discriminate between the distributions.
On the other hand, the minimal value of $D_n = 0.08$ and its associated p-value of $p = 0.91$ indicate that Clusters $C1$ and $C2$ yield the same distribution.
This contrasts the result of $k=5$ achieving the highest $D_n$ value and indicates that the clusters should be merged. 
Thus, exploring alternative clustering approaches and/or non-euclidean norms should be considered in future work.

%%%%%%%%%%%%%%%%%%%%%%%%%%%%%%%%%

When applying the junction/non-junction classification rule defined by us (human-centric, top-down), we obtain the ECDF of \Cref{fig:manual_classification}. 
In this case, the distribution of non-junction scenarios indicates that people accept reduced distances when not being in junctions.
We assume that this is partly due to the limited space available on the sidewalk.

Finally, we can compare the KS-statistics of \Cref{fig:ecdf_clustering} and \ref{fig:manual_classification}.
This shows that the discriminatory power of using the automatically generated clusters is compared to using the manual definition.
%Thus, the automatically generated clusters hold increased discriminatory power compared to the human-centric factor of junction/non-junction.
This supports our claim that a data-driven, bottom-up approach is necessary to enable a fine-grained analysis of robot-pedestrian interactions.

%% file: sections/04_future_work.tex
\section{Future Work}
\label{sec:future_work}

% \begin{itemize}
    
%     \item Es werden mehr datensätze und deutlich mehr feature gebraucht, um die Signifikanz zu verbessern
%     \item Bemerkenswert ist dabei, dass die geschätzte mittlere Geschwindigkeit der Passanten zu einem großen Anteil auch nahe null liegt. Menschen warteten also an bestimmten Stellen, während der Roboter vorbei fuhr. Weitere Betrachtungen zu den Interaktionsmustern sollten diesen Aspekt unbedingt aufgreifen und differenziertere Bilder der Interaktionmaneuver entwickeln.
%     \item Der context sollte mehr betrachtung finden. z.B. ist der Abstand mit Gehwegbreite korreliert oder sowas.
%     \item Wir sollten auch untersuchen, ob wenn wir die punkte nicht einzeln betrachten sondern per fahrt, sich bestimmte muster zeigen. (ich erwarte das ein überholmanöver immer gleiche verläufe der nähe hat bzw sich im extend unterscheided jenachdem wie schnell eine person überholt z.B.)
%     \item Dauert ein überholmanöver je nach winkel länger/kürzer?
%     \item sind leute auf die straße ausgewichen? 
% \end{itemize}

%Deploying mobile robots in urban environments promises a wide range of applications but entails conflicting interactions with other users of the infrastructure. 
%For increasing the social acceptance, interaction patterns need to be studied and the robot's behavior need to be adjusted correspondingly.

In this work-in-progress study we argue that a data-driven, bottom-up approach is required to analyze the interactions from a robot's perspective.
It provides not only fine-grained insights into the interactions but simultaneously defines approaches to make robots aware about them.

%The proposed pipeline for exploring robot-pedestrian interactions was evaluated using the RoboTraces dataset. 
%Analyzing 800 interactions recorded in the city center of Freiberg, Germany, we showed that the contextual factor of whether entities are close to junctions changes the minimal distance kept between them. 
%By applying unsupervised learning, we could additionally show the increased discriminatory power while simultaneously identified necessary future work to improve the clustering. 

% As such, the next steps are to evaluate the effects of alternative approaches in data preparation (add more features, evaluate normalization strategies, 

Given the complexity of interactions and the multitude of variables involved, it is imperative to expand our dataset for more comprehensive evaluations. Continued data collection in the same environment will enrich our analysis, allowing for a deeper examination of the factors influencing robot-pedestrian interactions. A particular focus will be on scenarios where humans wait for the robot to pass -- investigating whether their actions stem from necessity or curiosity, and how these interactions evolve as the local population becomes accustomed to the robot's presence.

Further research will also explore whether the distance between robots and humans correlates more significantly with certain contexts, such as the width of sidewalks. Analyzing the duration of overtaking maneuvers in relation to collision angles will provide insights into how spatial dynamics affect interaction patterns.